\documentclass{article}

\usepackage{arxiv}

\usepackage[utf8]{inputenc} 
\usepackage[T1]{fontenc}    
\usepackage{hyperref}       
\usepackage{url}            
\usepackage{amsfonts}       
\usepackage{nicefrac}       
\usepackage{microtype}      
\usepackage{lipsum}
\usepackage{graphicx}
\usepackage{amsmath}
\usepackage{bbm}
\usepackage{booktabs}
\usepackage{wrapfig}
\usepackage{multirow}
\usepackage[capitalise]{cleveref}
\usepackage{algorithm,algpseudocode}
\usepackage[algo2e]{algorithm2e} 
\usepackage{amssymb}
\usepackage{amsfonts}
\usepackage{diagbox}
\usepackage{color}
\usepackage{bm}
\usepackage{adjustbox}
\usepackage[numbers]{natbib}
\usepackage{tikz}
\usepackage{caption}
\usetikzlibrary{bayesnet}
\usetikzlibrary{arrows}
\usetikzlibrary{backgrounds}
\graphicspath{ {./images/} }

\usepackage{caption}
\usepackage{subcaption}
\usepackage{multirow}
\usepackage{float}
\newcommand{\tabincell}[2]{\begin{tabular}{@{}#1@{}}#2\end{tabular}}

\title{De-biased Representation Learning for Fairness with Unreliable Labels}

\author{
 Yixuan Zhang \\
  Data Science Institute\\
  University of Technology Sydney\\
  \texttt{yixuan.zhang@student.edu.au} \\
   \And
 Feng Zhou \\
  Department of Computer Science and Technology\\
  Tsinghua University\\
  \texttt{zhoufeng6288@tsinghua.edu.cn} \\
  \And
 Zhidong Li \\
  Data Science Institute\\
  University of Technology Sydney\\
  \texttt{zhidong.li@uts.edu.au} \\
    \And
 Yang Wang\\
  Data Science Institute\\
  University of Technology Sydney\\
  \texttt{yang.wang@uts.edu.au} \\
    \And
 Fang Chen \\
  Data Science Institute\\
  University of Technology Sydney\\
  \texttt{fang.chen@uts.edu.au} \\
}

\begin{document}
\maketitle
\begin{abstract}
Removing bias while keeping all task-relevant information is challenging for fair representation learning methods since they would yield random or degenerate representations w.r.t. labels when the sensitive attributes correlate with labels. Existing works proposed to inject the label information into the learning procedure to overcome such issues. However, the assumption that the observed labels are clean is not always met. In fact, label bias is acknowledged as the primary source inducing discrimination. In other words, the fair pre-processing methods ignore the discrimination encoded in the labels either during the learning procedure or the evaluation stage. This contradiction puts a question mark on the fairness of the learned representations. To circumvent this issue, we explore the following question: \emph{Can we learn fair representations predictable to latent ideal fair labels given only access to unreliable labels?} In this work, we propose a \textbf{D}e-\textbf{B}iased \textbf{R}epresentation Learning for \textbf{F}airness (DBRF) framework which disentangles the sensitive information from non-sensitive attributes whilst keeping the learned representations predictable to ideal fair labels rather than observed biased ones. We formulate the de-biased learning framework through information-theoretic concepts such as mutual information and information bottleneck. The core concept is that DBRF advocates not to use unreliable labels for supervision when sensitive information benefits the prediction of unreliable labels. Experiment results over both synthetic and real-world data demonstrate that DBRF effectively learns de-biased representations towards ideal labels. 
\end{abstract}


\section{Introduction}
\label{introduction}
Many empirical findings have shown that discrimination towards certain demographic groups exists in contemporary machine learning systems~\citep{pmlr-v81-buolamwini18a,bird2016exploring} in the high-stakes domains, such as finance~\citep{credit_example}, medicine~\citep{medical} and law~\citep{compas}. Label bias, which is modeled as the ideal (fair) labels are flipped systematically for individuals belonging to particular demographic groups, is in fact, acknowledged as the primary source inducing discrimination~\citep{unlocking_fairness}. For example, in the recidivism prediction tasks, when participants are provided with biased information, the predicted labels usually are biased against the protected racial group~\citep{human_bias_compas}. Another example is in recruitment, which has been shown that employers will make evaluations based on the applicant's name since it perceives the ethnic origin of the applicant~\citep{employer}, and the decisions are usually in favor of particular demographic groups. When decisions are made algorithmically based on such data, it will hamper accuracy and fairness, which brings harm to both society and individuals. 

Due to ethical imperatives, fairness-aware learning experienced a surge of advances in recent years. Among all these equitable and robust methods, fair representation learning is a promising approach and attractive due to its flexibility in relaxing the limitations of underlying learning algorithms for downstream tasks. Existing fair representation learning methods~\citep{2015-fvae,DBLP:flex,icml_2013,NIPS2017_optimised_preprocessing,2016_lum} usually try to remove bias by isolating the representations from explicit sensitive attributes in the training process and then obtain the so-called debiased representations. 

Nevertheless, how to ensure the representations are debiased as well as informative is still a challenging problem. First, removing all discriminative information while keeping all task-relevant information is hard since fair representation learning as a pre-processing approach usually does not work with label information. If the sensitive attributes correlate with the label, this will yield random or degenerate representations w.r.t. the labels~\citep{2015-fvae}. To this end, \citet{2015-fvae} and \citet{NIPS2017_optimised_preprocessing} proposed to incorporate label information into the representation learning procedure to overcome such issues. However, here comes the second challenge, for previous works requiring label information for supervision, when the observed labels are corrupted against specific demographic groups, the debiased representation will be contaminated again by the sensitive information leaked in the unreliable labels. Moreover, for methods that do not inject label information in the training process, the labels are still required for supervising downstream tasks. Therefore, when learning with unreliable labels, the debiased representations cannot guarantee the downstream classifiers to make fair predictions. 

To circumvent these issues, we explore the following question: \emph{Can we make the representations predictable to ideal labels when we only have access to unreliable labels?} Our work is motivated by the disentanglement learning~\citep{pmlr-v80-kim18b,Higgins2017betaVAELB} and information bottleneck~\citep{dai2018compressing,burgess2018understanding}. We propose a framework, namely \textbf{D}e-\textbf{B}iased \textbf{R}epresentation Learning for \textbf{F}airness (DBRF). As shown in~\cref{fig:model}, we consider three aspects to lead the learning process toward ideal labels (fair but latent). The first aspect is that we disentangle the latent representation into biased and fair representations. The second aspect is that we use the information bottleneck and mutual information to control the impact of these two components. The third aspect is that we utilize the biased representations to recognize which instances are likely to be discriminated by the label bias. Intuitively, if the biased representation can predict the labels well, we regard this case with a higher chance of being discriminated against and vice versa. 

The major contributions of our work are summarized as: \textbf{(1)} To the best of our knowledge, under the problem setting of learning with unreliable labels, we propose the first fair representation learning framework attempting to recover the ideal labels. \textbf{(2)} We present a flexible end-to-end framework that can either directly learn the ideal labels without separating the learning process into two stages or extract the fair representations which can be used for downstream classifiers. \textbf{(3)} We empirically show that biased labels are adverse to both accuracy and fairness, even when the learned representations remove the bias encoded in input attributes. With unreliable labels, experimental results demonstrate that DBRF effectively learns de-biased representations towards ideal labels.

\section{Preliminary}
\label{sec:preliminary}
\textbf{Fair Representation Learning} In fair classification, we consider triplet of random variables $(x,y,a)$ with sample space $\Omega = \mathcal{X} \times \mathcal{Y} \times \mathcal{A}$, where $x$ denotes the non-sensitive attributes ($x$ may also contain implicit sensitive information), $y$ represents observed unreliable labels contaminated by the biasing information, and $a$ are sensitive attributes. In a binary sensitive attribute setting, $a=1$ denotes the protected group and $a=0$ the privileged group. Fair representation learning aims to minimize the correlation between the learned representation $z$ and $a$ (i.e., achieve $z \perp a$). Then, the prediction $\hat{y}$ from a labeling function based on $z$ is independent of $a$, which is considered as a fair label. 

\textbf{Group Fairness} In order to measure the fairness violation, many existing works applied the notion of group fairness. The basic idea of group fairness is that if we define a demographic group we want to protect, then we ask for parity between the protected group and all other groups with different statistical measures. In this work, we use the demographic parity distance metric ($\Delta_{\text{DP}}$) as the measure for fairness violation (other measure criteria, such as difference of equal opportunity (DEO)~\citep{equal_opportunity} or p\%~\citep{biddle}, can also be applied). The definition of $\Delta_{\text{DP}}$ is given by: 
\begin{equation}
\label{eq:dp}
    \Delta_{\text{DP}} = |\mathbb{E}(\hat{y}=1\mid a=1) - \mathbb{E}(\hat{y}=1\mid a=0)|,
\end{equation}
which states that the prediction should be independent of sensitive attributes and we achieve demographic parity when $\Delta_{\text{DP}}=0$.

\textbf{Disentanglement VAE} The aim of a VAE~\citep{kingma2014autoencoding} is to learn the latent representations $z$ and use them to reconstruct the input. The vanilla VAE is implemented by adding a sampling layer linking the encoder and decoder. The samples are usually drawn from an isotropic Gaussian distribution. Since directly maximizing the data log-likelihood is difficult, the standard VAE objective function uses the evidence lower bound (ELBO) instead. In standard VAE, the regularization (KL) term tries to make the distribution of outputs from the encoder as close to the prior of $p(z)$ as possible. Therefore, in order to control the model's adherence to $p(z)$, $\beta$-VAE~\citep{Higgins2017betaVAELB} is then proposed by simply adding a hyperparameter $\beta$ times the KL term. When $\beta$ increases, it encourages the output from the encoder to be disentangled. To allow the latent representation to be further disentangled at each dimension, \citet{pmlr-v80-kim18b} proposed FactorVAE by adding an additional weighted total correlation term: $\gamma \text{KL}[q(z)||\prod_j q(z_j)].$ In this work, we employ FactorVAE to disentangle the fair representations and biased representations.

\textbf{Information Bottleneck}
The information bottleneck (IB) technique is introduced by~\citet{Tishby99theinformation}. In general, the objective function, measuring how well $y$ is predicted from the compressed encoding $z$, can be expressed in the following format: $\min I(x;z) - \beta I(z;y).$, where $I(x;z)$ is the mutual information between original input $x$ and the compression term $z$, and $I(z;y)$ is the mutual information between the compression term $z$ and output $y$. By optimizing the difference between two mutual information, such that the compressed encoding learned from the training set is highly informative for prediction, we can find the best trade-off between accuracy and complexity. IB has been used in many applications such as regularization~\citep{alemi2019deep}, understanding of $\beta$-VAE~\citep{burgess2018understanding} and compression for deep neural networks~\citep{dai2018compressing}. In this work, we utilize IB to penalize the leakage of biased information to the labels.

\section{Method}

In this section, we illustrate our design for DBRF. In \cref{sec:problem_formulation}, we introduce the problem formulation and motivations by proposing two essential questions. In \cref{sec:first_q,sec:sec_q}, we discuss how we solve these questions; based on that, we propose DBRF. In \cref{sec:opt}, we introduce the optimization for DBRF.

\subsection{Problem Formulation}
\label{sec:problem_formulation}

Learning a fair representation $z$ with unreliable labels is a challenging task. In this work, apart from learning the fair representation $z$, we try to learn another latent representation $b$, which preserves all sensitive information that can infer $a$. To simplify the notations throughout this paper, we do not specify the format of sensitive attribute $a$, and it can be either binary or multi-attribute (e.g., race v.s. race$\land$gender). We assume the unreliable observed labels $y$ are corrupted with the probability of $\rho_0 = p(y=0 \mid y_m =1, a=1)$ and $\rho_1 = p(y=1 \mid y_m = 0, a=0)$ for individuals belonging to protected group and privileged group respectively where $y_m$ denotes the ideal (fair but latent) labels. To learn $z$ which is predictable to $y_m$ when only $y$ is observed, we introduce an auxiliary variable $r_m$ which is a proxy label of $y_m$ associating with the representation $z$ drawn from the decoder, and we use $\hat{y}_m = \sigma(r_m)$ to represent the learned ideal labels, where $\sigma$ is the sigmoid function.

\begin{figure}[t]
  \centering
  \includegraphics[width=0.99\linewidth]{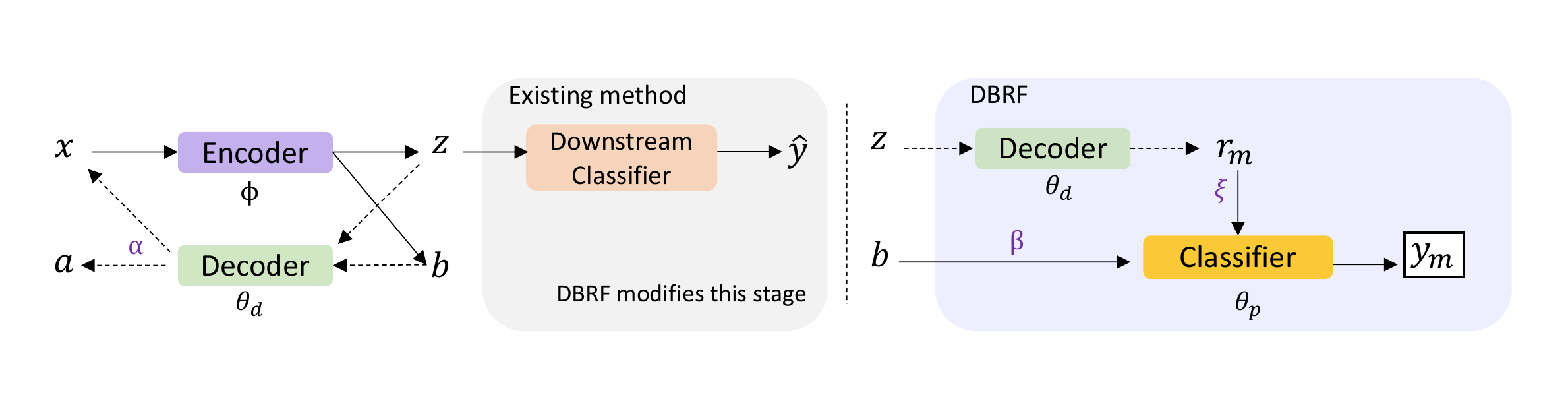}
  \caption{Framework of DBRF. $x$, $a$ and $y$ are observations, $z$ and $b$ are representations learned from $x$, and we require $z \perp b$ in the latent space. $r_m$ is decoded from $z$. $\xi$ and $\beta$ are two hyperparameters that control the influence from $r_m$ to $y$ and $b$ to $y$, respectively. The model parameters are written below the corresponding model. DBRF (purple shading) is different from existing method (gray shading) by injecting $b$, $r_m$ and label information to infer ideal label $y_m$.}\label{fig:model}
\vskip -0.2in
\end{figure}

A sketch of the DBRF framework is shown in~\cref{fig:model}. Specifically, we use disentanglement VAE to guarantee the obtained fair representation $z$ and biased representation $b$ are independent of each other and then apply the IB to control the bias information leakage and infer the ideal labels $y_m$. To start with our proposed method, we first emphasize some key motivations in the framework by answering the following questions: (1) How can we guarantee the independence between the fair representation $z$ and biased representation $b$ in the latent space? (2) How to ensure the output $\hat{y}_m$, which is obtained from $z$, is ideal? In the following sections, we discuss how we solve these two questions in detail. 

\subsection{Disentanglement Between $z$ and $b$}
\label{sec:first_q}

The first question has been investigated in some recent works of fair representation learning~\citep{DBLP:flex,2015-fvae} either by applying KL regularization or using Maximum Mean Discrepancy on representations to force $z$ to discard the information about $a$ in the latent space. Here, we utilize the FactorVAE~\citep{kim2019disentangling} to de-correlate $z$ and $b$. 

A similar method has been proposed by \citet{DBLP:flex} where the difference is that in our framework, for multiple sensitive attributes, we do not care if the corresponding dimension of $b$ can represent and align with each dimension of $a$. We only focus on disentangling the sensitive information from $z$ and keeping $z$ as fair as possible. Based on that, we encourage $I(z;b)$ to be as low as possible. Therefore, the objective function of this part can be formulated in a form that is similar to the objective function of FactorVAE: 
\begin{equation}
\begin{aligned}
\label{eq:loss1_vae}
  \mathcal{L}_{\text{DBVAE}} &= \mathbb{E}_{q_\phi(z,b\mid x)}[\log p_{\theta_d} (x\mid z,b)+\alpha \log p_{\theta_d} (a\mid b)] 
  - \text{KL}[q_\phi(z, b \mid x)||p(z, b)] \\& -  \gamma \text{KL}[q_\phi(z,b)||q(z)q(b)],
\end{aligned}
\end{equation}
where $q_\phi(z,b\mid x)$ is the encoder, $p_{\theta_d}(a\mid b)$ and $p_{\theta_d}(x\mid z,b)$ are two separate decoders. The encoder tries to encode $x$ into the fair representation $z$ and biased representation $b$. On the contrary, the decoder $p_{\theta_d}(a\mid b)$ tries to decode biased representation $b$ back to the sensitive attribute $a$ and $p_{\theta_d}(x\mid z,b)$ tries to reconstruct $x$ based on both biased and fair representations. $\alpha$ and $\gamma$ in \cref{eq:loss1_vae} are two hyperparameters: $\alpha$ controls the degree of how $b$ can infer back to $a$, and $\gamma$ controls the degree of disentanglement between $b$ and $z$. In FactorVAE, the fatorization in additional weighted total correlation term $\gamma\text{KL}[q(z)||\prod_jq(z_j)]$ allows the latent representations to be further disentangled, i.e., $z_i$ correlates with $z_j$ iff. $i=j$. Here, we consider the latent representation of a distribution $q_\phi(z,b)$ is factorized into $q(z)$ and $q(b)$, and the term $\text{KL}[q_\phi(z,b)||q(z)q(b)]$ is just the $I(z;b)$. To compute this weighted correlation term, we apply the similar approach that is introduced in FactorVAE. This term is estimated by using a discriminator that distinguishes `real samples' and `fake samples', and then applying the density-ratio trick for approximation (see \cref{sec:app_implementation}). 

\subsection{Guarantee of $y_m$ Being Ideal}
\label{sec:sec_q}

For the second question, we formulate the problem using mutual information and information bottleneck~\citep{Tishby99theinformation}. The idea is to let $z$ be highly informative to predict $y_m$. Concurrently, we discourage $y_m$ to be similar to $y$ for samples with a high chance of being unfair against certain demographic groups. 

\textbf{Objectives.} In general, to achieve this goal, we need to satisfy several conditions: (1) the representations $z$ and $b$ are not `copied' from $x$ directly (i.e., do not overfit $x$); (2) $r_m$, as decoded from $z$, should be independent of $b$ as much as possible; (3) $r_m$ and $b$ are both predictable to $y$ to some extent. For discriminated samples, $b$ is more powerful than $r_m$ for predicting $y$ and vice versa. Based on the above conditions, we formulate the IB objective as: 
\begin{equation}
\label{eq:lib}
    \mathcal{L}_{\text{IB}} = \lambda I(x;z,b) + I(b;r_m) - \beta I(b;y) - \xi I(r_m;y),
\end{equation}
where $\lambda \in [0,1]$, $\beta \in [0,1]$ and $\xi \in [0,1]$ are the hyperparameters.  $I(x;z,b)$ is the compression term between $x$ and the joint variables $z$ and $b$. We use $\lambda$ to control the compression degree. We need to point out that, we learn the biased representation $b$ from $x$ and require it can reconstruct $a$ well to remove the bias encoded in the non-sensitive attributes (the reconstruction loss in \cref{eq:loss1_vae}), so we focus on the compression from $x$ into $z$ and $b$. $I(b;r_m)$ is the term indicating the dependency between biased representation $b$ and the proxy label $r_m$ that is decoded by $z$. We require $r_m$ and $b$ to be independent of each other to guarantee the learned soft label is clean, i.e., $r_m \perp b$. $I(b;y)$ indicates how $b$ potentially contributes the biased information to predict unreliable label $y$. We use $\beta$ to penalize its impact. If the ratio of unreliable labels is high, we increase $\beta$ and vice verse. $I(r_m;y)$ indicates how well the learned latent proxy label $r_m$ predicts the observed label $y$. We use $\xi$ to adjust the information from $r_m$ to $y$.

\textbf{Supervision for $\bm{r_m}$.} To further encourage both $r_m$ and $z$ to be fair, we add another supervision constraint between $f_{\theta_p}(z)$ and $y$, where $f$ is a classifier (e.g., logistic regression, SVM or MLP) parametrized by $\theta_p$. We use the confidence score of $y$ predicted from $b$ as the weights for the supervision loss, i.e., the more confident we predict $y$ using biased representation, the more chance such instance is discriminated by the sensitive information, and consequently, we need to penalize the impact from such instance to the loss. Based on this, we define the supervision loss as:
\begin{equation}
\label{eq:lp}
    \mathcal{L}_{\text{p}} = w_b\ell( f_{\theta_p}(z),y), 
\end{equation}
where $w_b = 1-p(y \mid b)$ and $\ell$ is the cross entropy loss.

\subsection{Optimization for DBRF}
\label{sec:opt}
Combining \cref{eq:loss1_vae,eq:lib,eq:lp}, we obtain the objective function for DBRF:
\begin{equation}
\label{eq:obj_intial}
    \mathcal{L_{\text{DBRF}}} = -\mathcal{L}_{\text{DBVAE}}+\mathcal{L}_{\text{IB}}+ \mathcal{L}_{\text{p}}. 
\end{equation}
Clearly, in \cref{eq:obj_intial}, the optimization for $\mathcal{L}_{\text{DBVAE}}$ and $\mathcal{L}_{p}$ is straightforward. However, $\mathcal{L}_{\text{IB}}$ is an intractable function and we cannot optimize it directly. In this section, we simplify the four mutual information terms in $\mathcal{L}_{\text{IB}}$~\cref{eq:lib}, and then present a tractable equivalent loss of $\mathcal{L}_{\text{DBRF}}$. 

\textbf{Simplify $\mathcal{L}_\text{IB}$.} We first express all the $I(\cdot;\cdot)$ terms in \cref{eq:lib} in the form of conditional entropy based on the properties of mutual information (See \cref{sec:app_mi_upper_bound} for more details). The four terms in \cref{eq:lib} can be rewritten as:
\begin{equation}
\label{eq:four_terms}
\begin{aligned}
    I(x;z,b)&=\text{KL}[q(z,b\mid x)||p(z,b)], \   \    \    \    I(b;r_m) = I(r_m;y)-H(y\mid b)+H(y\mid b,r_m), \\
    I(r_m;y) &= H(y) - H(y\mid r_m),\   \    \    \     \   \    \    \       \   \   
    I(b;y) = H(y) - H(y\mid b),
\end{aligned}
\end{equation}
where $H(\cdot)$ indicates the marginal entropy and $H(\cdot \mid \cdot)$ indicates the conditional entropy. Substituting \cref{eq:four_terms} into \cref{eq:lib}, we obtain: 
\begin{equation}
\begin{aligned}
\label{eq:lmi}
  \mathcal{L}_{\text{IB}} &= \lambda\text{KL}[q(z,b\mid x)||p(z,b)] + \xi H(y\mid r_m) + \beta H(y\mid b) - (\xi+\beta)H(y),\\
  &= \lambda\text{KL}[q(z,b\mid x)||p(z,b)] + 
  \xi H(y\mid r_m) + \beta H(y\mid b) - C. 
\end{aligned}
\end{equation}
The first term is same as the first KL term in $\mathcal{L}_{\text{DBVAE}}$ of \cref{eq:loss1_vae} which is tractable. $H(y)$ is the entropy of the observation so it is a constant w.r.t. model parameters and we rewrite $(\xi+\beta)H(y)$ as $C\ge 0$. We drop the constant $C$ and define:
\begin{equation}
\label{eq:umi_loss}
    \mathcal{L}_{\text{UMI}} := \xi H(y\mid r_m) + \beta H(y\mid b)
\end{equation}
 
Combining \cref{eq:loss1_vae,eq:lp,eq:umi_loss}, we obtain a tractable equivalent objective function: 
\begin{equation}
\begin{aligned}
\label{eq:objective_function}
    \mathcal{\tilde{L}_{\text{DBRF}}} &= -\mathcal{L}_{\text{DBVAE}}+\mathcal{L}_{\text{UMI}}+ \mathcal{L}_{\text{p}} \\
    &= -\mathbb{E}_{q_\phi(z,b\mid x)}[\log p_{\theta_d} (x\mid z,b)+\alpha \log p_{\theta_d} (a\mid b)] + (1+\lambda)\text{KL}[q_\phi(z, b\mid x)||p(z, b)]\\
    &+  \gamma \text{KL}[q_\phi(z,b)||q(z)q(b)] +\xi H(y\mid r_m)+\beta H(y\mid b) +w_b \ell(f_{\theta_p}(z),y). 
\end{aligned}
\end{equation}

$\mathcal{\tilde{L}_{\text{DBRF}}}$ and $\mathcal{L_{\text{DBRF}}}$ differ only by the constant $C$. It is worth noting that a factor $(1+\lambda)$ is used to scale the KL term between $q_\phi(z, b \mid x)$ and $p(z,b)$. The $\lambda$ term is derived from the compression term $I(x;z,b)$ as we mentioned above, and the factor $(1+\lambda)$ works in the same manner as the hyperparameter $\beta$ in the $\beta$-VAE does, which is used to control the model's adherence to $p(z,b)$. When $\lambda=0$, this is equivalent to a standard VAE.

\textbf{Insight of DBRF objective function.} By simplifying $\mathcal{L}_{\text{IB}}$ in \cref{eq:lib}, we obtain some insights to explain our proposed framework more in-depth. If we check each term in \cref{eq:umi_loss}, we can find that the first term $H(y\mid r_m)$ is exactly the cross-entropy between $y$ and the prediction of $y$ using $r_m$. Similarly, the second term $H(y \mid b)$ is the cross-entropy between $y$ and the prediction of $y$ using the biased representation $b$, and this allows us to incorporate the cross-entropy loss directly and easy for optimization. In the case of individuals who have been labeled with bias, the prediction of $y$ from $b$ is supposed to be more `accurate' than that from $r_m$, which increases the impact of $b$ on the decision. In that case, we will increase $\beta$ to penalize $H(y\mid b)$ term for such discriminated samples. We will empirically show the regularization effects of these two terms in the experiment section.

\textbf{Training process.} In DBRF, we utilize the confidence score for the prediction of unreliable label $y$ using $b$ (i.e., $p(y\mid b)$) as the penalty for instances that are likely to be discriminated; however, how $\beta$ in $H(y\mid b)$ and $w_b$ work are different. We use $\beta$ to control the \emph{global} regularization impact on the learning process by using the biased information encoded in the data as a shortcut to predict. While $w_b$, obtained from optimizing $H(y\mid b)$, is a weight working on the \emph{individual} level to further penalize the instance that is more likely to be predicted by its sensitive information. Note that we use an auxiliary latent proxy label $r_m$, which is parameterized by $\theta_d$ (the decoder) and later influenced by both biased information and unbiased information through $H(y\mid r_m)$ and $H(y\mid b)$. Therefore, $r_m$ guarantees removing not only the bias encoded in the original attribute $x$ but also the bias contained in the unreliable labels. DBRF allows to output either the ideal label $y_m$ directly or the fair representation $z$ for downstream tasks. Since $r_m$ is decoded from $z$, the fair representation $z$ obtained from DBRF will also reduce the impact of label bias. We first obtain the latent representations $z$ and $b$ from the encoder, then we obtain $r_m$ from the decoder. After that, we use $z$, $r_m$ and $b$ to predict y, respectively. Note that the penalty weight $w_b = 1-p(y\mid b)$ is obtained in this step. By optimizing \cref{eq:objective_function}, we inject the debiased label information into the learning procedure towards $r_m$ and $z$, which aims to remove bias encoded both in the attributes and labels. 

\section{Related Work}
Recent fair representation learning approaches share a similar framework, which we have introduced in \cref{sec:preliminary}. \citet{2011_dwork} proposed an initial fairness representation learning framework. However, this method requires a distance metric used for similarity measures and only works with given data, which cannot be formulated as a generalization task. \citet{icml_2013} proposed an improved framework that still had limitations on representation since it used clustering for probabilistic representation mapping. Besides, the information of sensitive attributes may still have leakage issues in this method. Based on~\citet{icml_2013}, \citet{2015-fvae} proposed a method to tackle these limitations by using the framework of VAE~\citep{kingma2014autoencoding} for fair representation learning. In \citet{2015-fvae}, the authors injected label information into the model since the unsupervised method would give random representations to $y$ when $a$ and $y$ are correlated. The authors also presented a semi-supervised model to address the missing label problem. Moreover, they used Maximum Mean Discrepancy to avoid the leakage from $a$ when learning $z$. Nevertheless, the above methods all focused on the binary sensitive attribute. \citet{DBLP:flex} proposed an improved method that can deal with multi-attribute fair representation learning by introducing a factorized structure in the aggregate latent labels by utilizing the disentanglement VAE~\citep{burgess2018understanding,kim2019disentangling,chen2019isolating}. In~\citet{DBLP:flex}, $b$ and $z$ are disentangled from each other, and $b$ is further required to be independent on each dimension. Unlike the aforementioned works, our work focuses on another challenging problem: learning latent fair representations with unreliable labels.

\section{Experiment}
\label{sec:exp}
To evaluate the effectiveness and robustness of the proposed method, we compare it to several VAE-based fair representation learning baselines on both synthetic and real-world data under different label bias settings. Since we cannot observe the ideal labels in real-world data, we assume the observed data is clean and randomly flip the labels to create a biased version. In the following sections, we introduce our experimental settings, including datasets, baselines, and evaluation metrics.

\subsection{Datasets, Evaluation Metrics and Baselines}
We conduct experiments on one synthetic dataset and two real-world datasets, Adult and Compas. The statistics of all datasets are shown in \cref{tab:data_description}. We list the number of instances, the specified protected and privileged groups, as well as their corresponding number of instances. We use $\Delta_{\text{DP}}^*$ to denote the initial fairness violation under clean distribution. The detailed description for each dataset can be found in \cref{sec:app_dataset}. 

\begin{wraptable}{r}{0.57\columnwidth}\vspace{-.45cm}
\vskip -0.0in
\caption{Dataset description.}
\centering
\resizebox{\linewidth}{!}{%
\begin{tabular}{c|c|c|c|c} 
\toprule
Dataset & \tabincell{c}{\# of\\Instances} & \tabincell{c}{Protected/Privileged\\Groups} & \tabincell{c}{\# of\\Instances} & $\Delta_{\text{DP}}^*$ \\ 
\midrule
Synthetic & 10,800 & a=1/a=0 & 5150/5650 & 0.02 \\
\midrule
\multirow{2}{*}{Adult} & \multirow{2}{*}{30,717} & female/male & 10,067/20,650 & 0.20 \\
  &   & female \& black / rest & 1943/28,774 & 0.19 \\
\midrule
\multirow{2}{*}{Compas} & \multirow{2}{*}{5,554} & black/white & 2,874/2,680 & 0.15\\
 & & black \& male / rest & 2,848/2,706 & 0.14\\
\bottomrule
\end{tabular}
}
\label{tab:data_description}
\vskip -0.2in
\end{wraptable}

We compare our method with several VAE-based models, including the vanilla VAE~\citep{kingma2014autoencoding}, Flexibly Fair Variational Autoendoer (FFVAE)~\citep{DBLP:flex} and Fair Variational Autoencoder (FVAE)~\citep{2015-fvae}. We implement our proposed DBRF framework in two different versions: (i) DBRF*, as shown in~\cref{fig:model}, where we use $r_m$ from the decoder as predicted labels; (ii) DBRF+LR, where the labels are predicted from $z$ by employing a downstream classifier. In the experiment, we use logistic regression for illustration, but it can be extended to other classifiers (e.g., SVM or MLP) straightforwardly. For a fair comparison, we fix the autoencoder structure among the above models. The details of implementation can be found in \cref{sec:app_implementation}.

We use accuracy to measure the performance and $\Delta_{\text{DP}}$ to measure the fairness violation. It is worth noting that other group fairness notions can also be applied, and we listed the results under different fairness notions (see \cref{sec:app_other_fair}). A lower $\Delta_{\text{DP}}$ indicates a minor fairness violation. We split the data into 90\% train and 10\% test, and report the results of ten-fold experiments with random splits. We modify the training dataset with label bias in a similar manner with~\citet{unlocking_fairness}. We systematically flip the labels of individuals in the disadvantaged group from 1 to 0 with probability $\rho$ and flip the labels from 0 to 1 with $\rho$ for the privileged group (i.e., $\rho_0 = \rho_1 = \rho$). We test the performance of different methods under different settings of label bias with the flipping rate $\rho$ ranging from 0 to 0.45 while keeping the test dataset clean.

\subsection{Comparison Results}
\begin{figure}[ht]
\centering
\subfloat[Synthetic.\label{fig:1a}]{\includegraphics[width=0.195\textwidth]{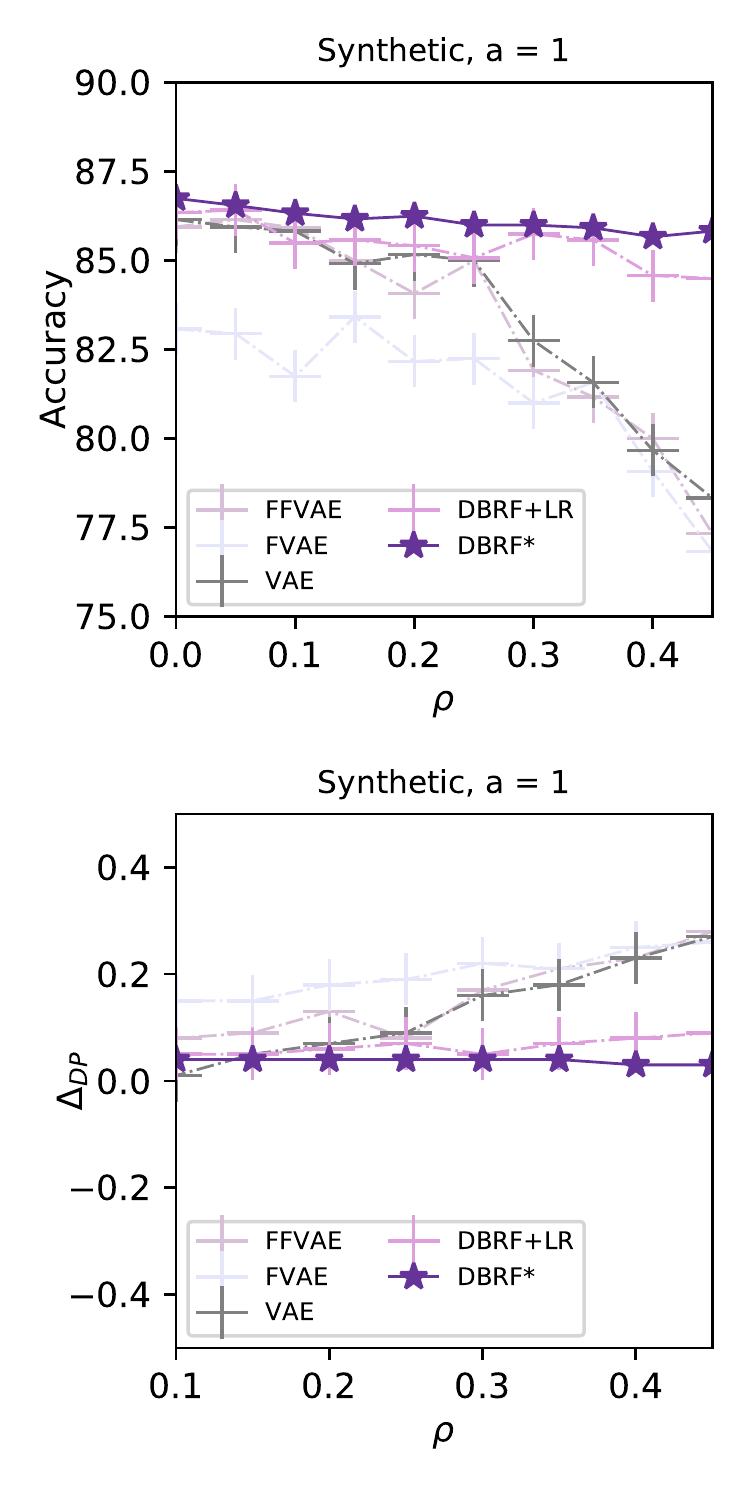}}
\subfloat[Adult.\label{fig:1b}] {\includegraphics[width=0.195\textwidth]{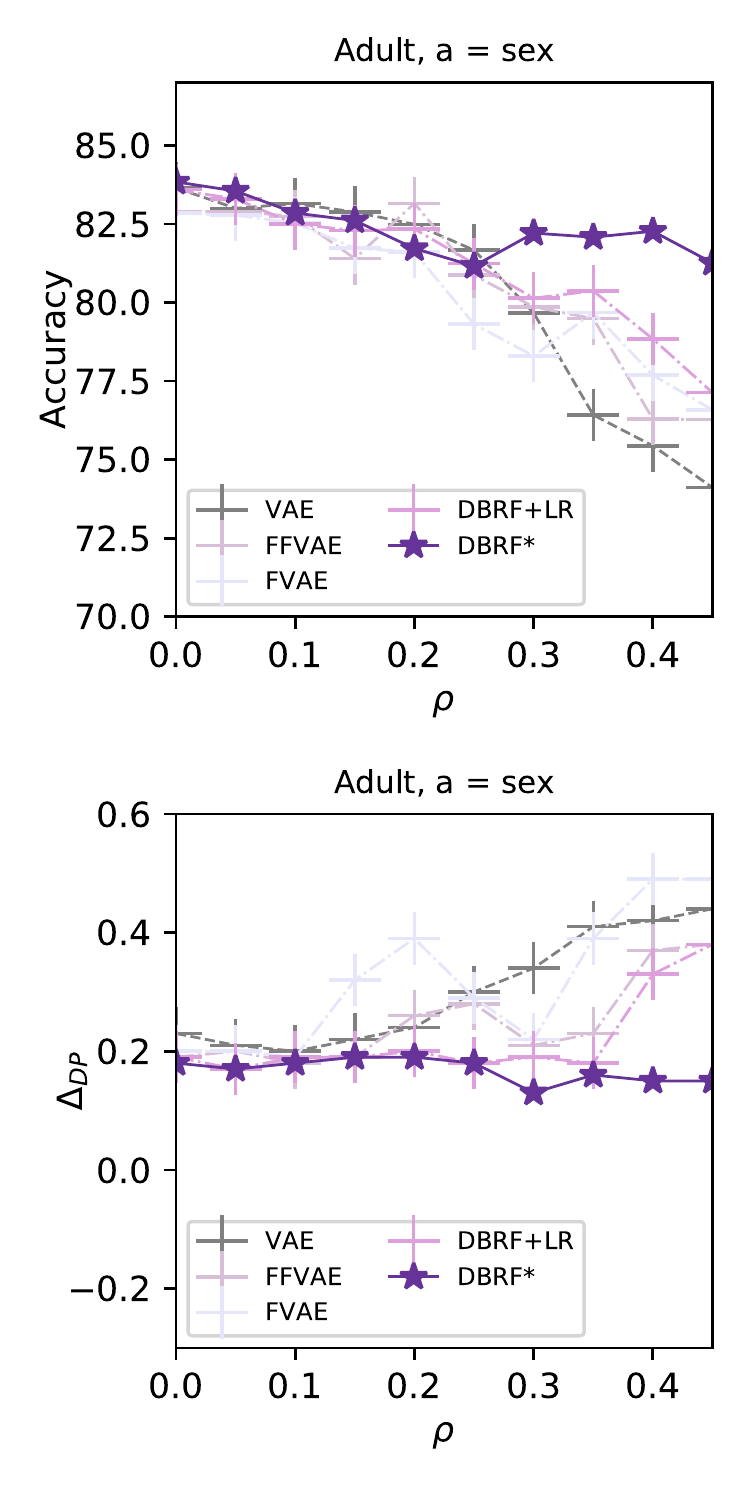}}
\subfloat[Adult (2d).\label{fig:1c}]{\includegraphics[width=0.195\textwidth]{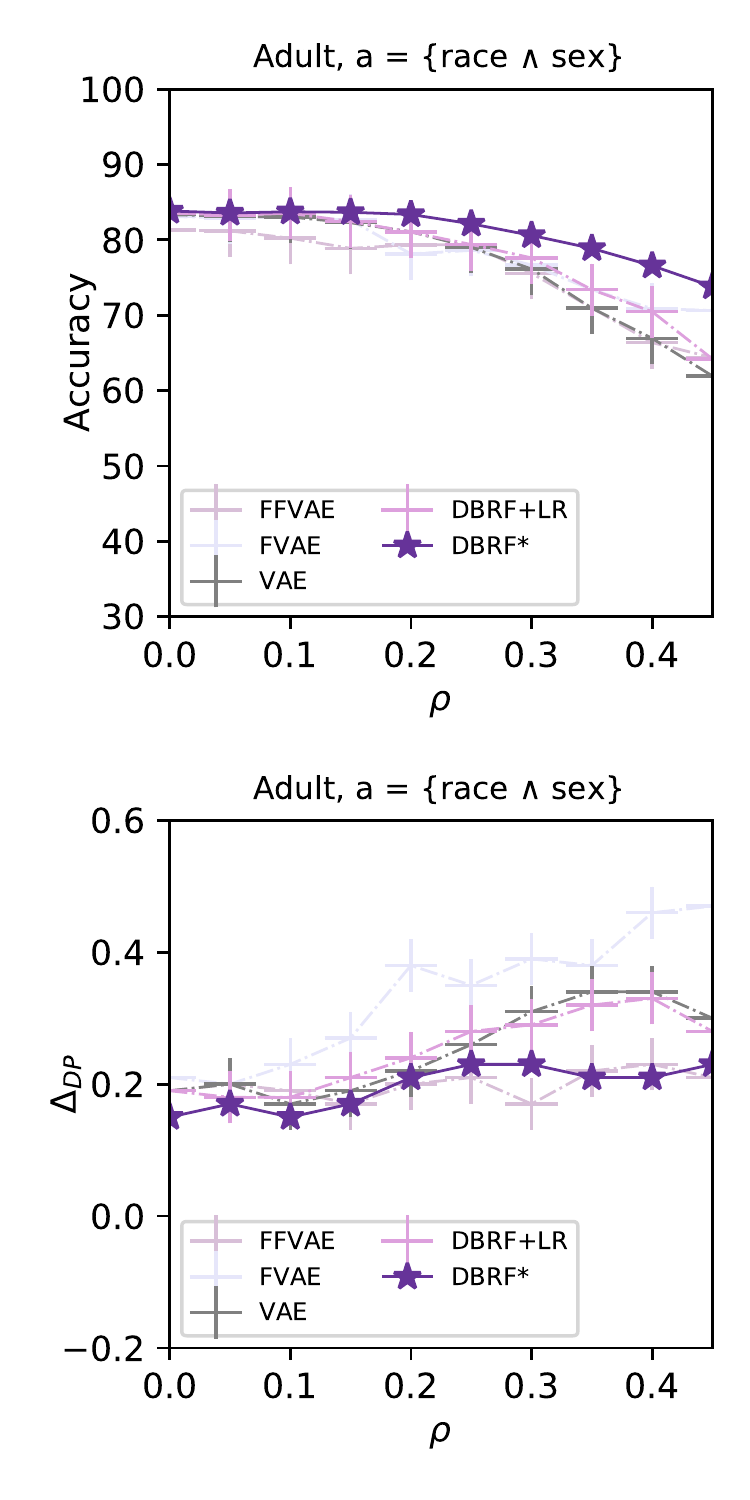}}
\subfloat[Compas.\label{fig:1d}]{\includegraphics[width=0.195\textwidth]{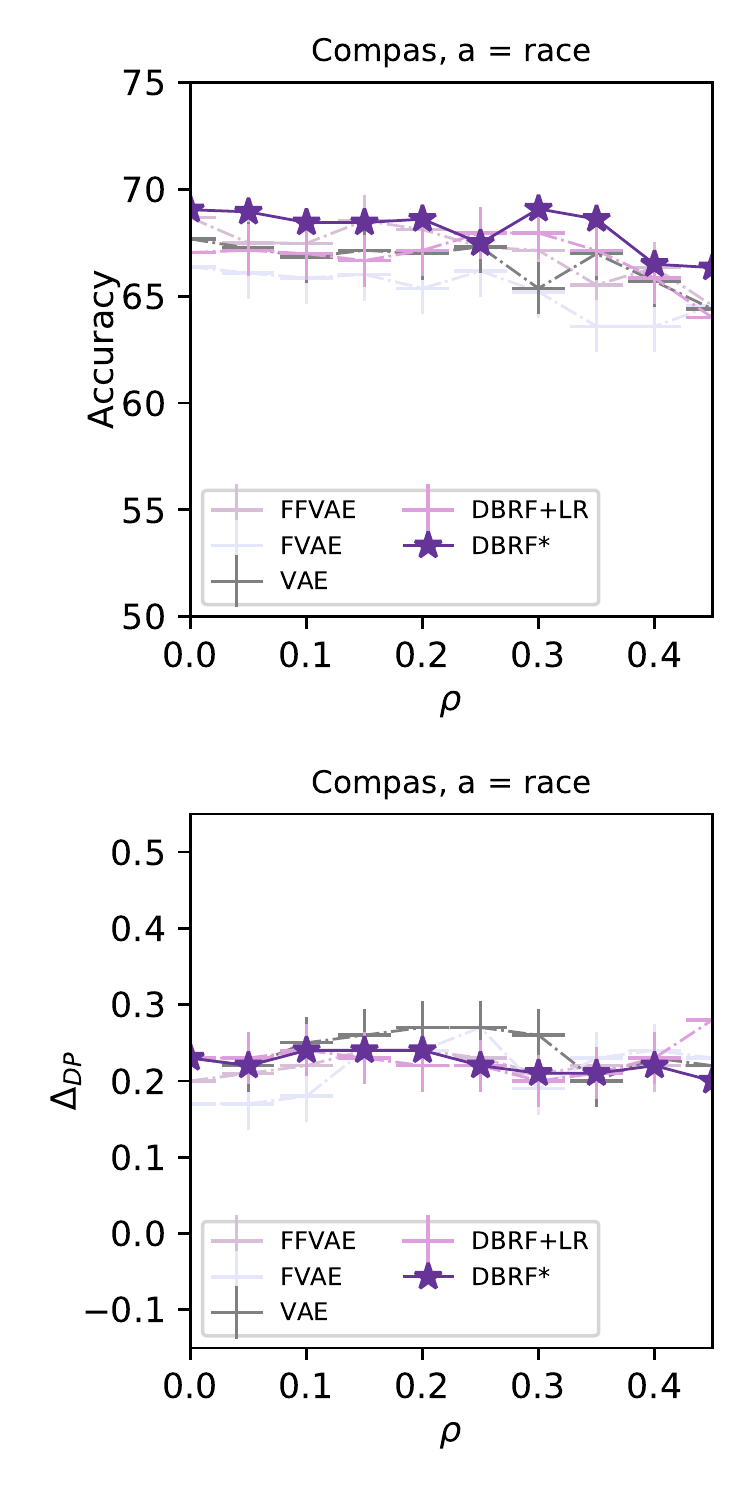}}
\subfloat[Compas (2d).\label{fig:1e}]{\includegraphics[width=0.195\textwidth]{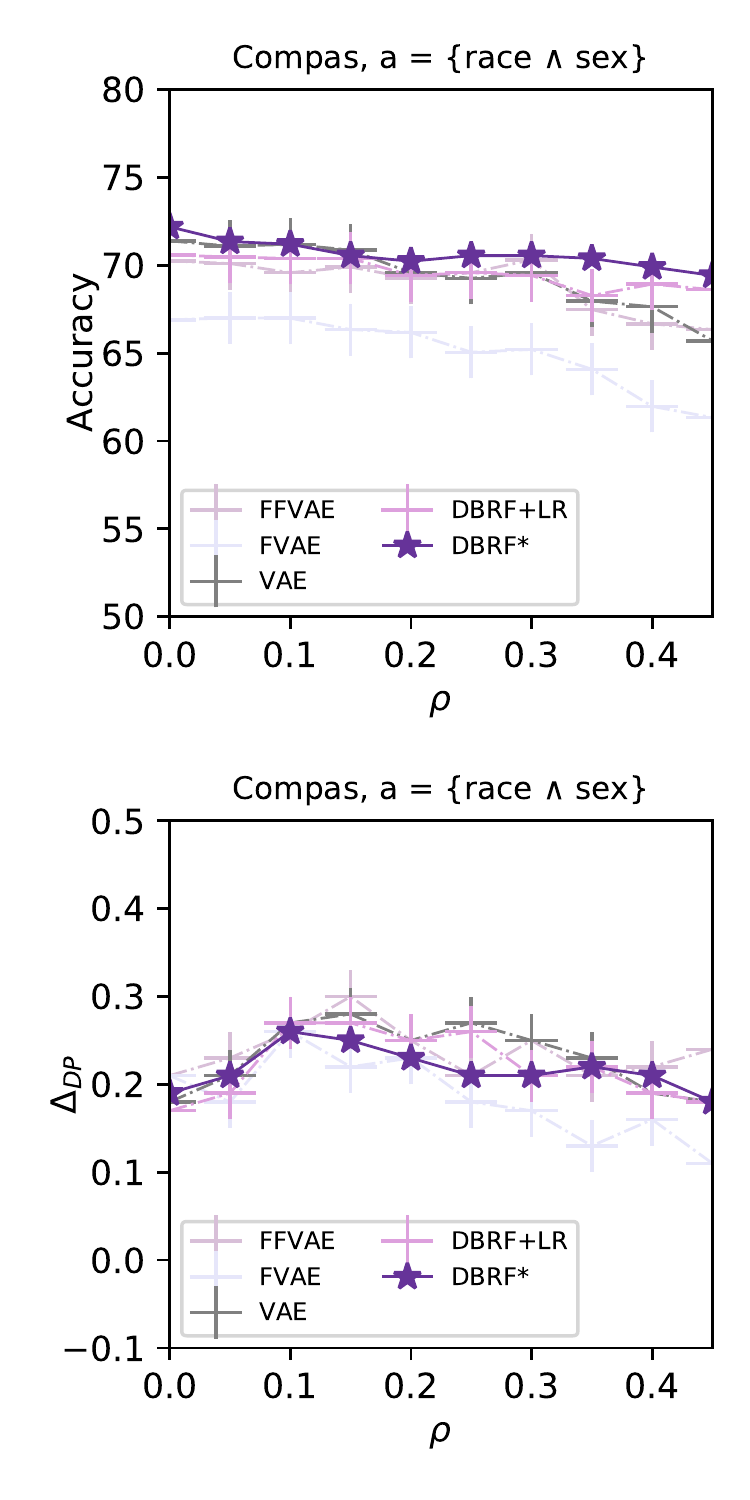}}
\caption{Accuracy and $\Delta_{\text{DP}}$ under different label bias settings. We start from clean data and incrementally add the bias from 0.5 to 4.5. For DBRF, when $\rho<0.3$, we fix $\alpha=1$, $\lambda=0.1$, $\xi=0.1$, and $\beta=0.1$ for the reported results. When $\rho >=0.35$, we increase $\beta$ to 0.5.}
\label{fig:acc_fairness}
\vskip -0.1in
\end{figure}

\textbf{Case 1: Binary sensitive attribute. }
The results are shown in \cref{fig:acc_fairness}. The prediction performance of our method generally outperforms other baselines with the increase of label bias w.r.t. both effectiveness and robustness. Overall, when the label bias increases to above 0.25, the accuracy of other methods starts to drop dramatically, and the fairness violation starts to increase, which demonstrates that no matter how fair the learned $z$ is, if we let $z$ be predictable to the biased label $y$, we still obtain the biased output. This conclusion can also be obtained by comparing DBRF* to DBRF+LR. Though the performance and fairness violation of DBRF+LR is generally better than other baselines, it is still worse than DBRF*, in which the predicted labels are directly obtained from $r_m$. We obtain the same trend in these three datasets. For the synthetic dataset, we find DBRF* has the overall highest accuracy with $\Delta_{\text{DP}}$ close to the $\Delta_{\text{DP}}^*$ under clean distribution. Also, we find the results of DBRF+LR are very close to DBRF* while DBRF+LR has a slightly higher fairness violation and lower accuracy. Other baselines work well when the bias amount is small, but when we increase the bias impact, they are not robust to the change of bias amount. For the Adult dataset, the accuracy of DBRF* is the highest among all other baselines, and at the same time, DBRF* has the closest $\Delta_{\text{DP}}$ to the $\Delta_{\text{DP}}^*$. When the label bias amount increases, the accuracy of DBRF* decreases slightly, but the performance of DBRF* is the most steady one compared to other baselines. For the Compas dataset, DBRF* has the highest accuracy. Though all methods have higher $\Delta_{\text{DP}}$ than $\Delta_{\text{DP}}^*$, DBRF* has the lowest $\Delta_{\text{DP}}$. Note that the $\rho$ starts from 0, so we can see that even on the "clean" data, our method still obtains the highest accuracy with the lowest fairness violation in most cases. Overall, DBRF is superior to the other baseline methods for learning the de-biased representations under different label bias amounts.

\textbf{Case 2: Multiple sensitive attributes. } In this section, we conduct experiments on Adult and Compas datasets under the setting that we have 2-dimensional sensitive attributes, which are specified in \cref{tab:data_description}. The results are shown in \cref{fig:acc_fairness}. For the Adult dataset, the task is more difficult than the binary sensitive attribute setting since the number of instances among the protected and privileged groups is more imbalanced than in the binary sensitive attribute setting. We can clearly see that the accuracy and fairness of the Adult dataset for all baselines perform worse than the binary setting. Even so, DBRF* still has the highest accuracy compared to the other baselines. Also, $\Delta_{\text{DP}}$ of DBRF* is the lowest and closest to $\Delta^*_{\text{DP}}$. We notice that the accuracy of DBRF+LR is close to DBRF* but with higher fairness violations. For the Compas dataset, the number of instances for the protected and privileged groups is balanced as in the binary sensitive attribute case. In this case, DBRF* has the highest accuracy, and the performance of DBRF+LR is very close to DBRF*. It is worth noting that, though FVAE has the lowest $\Delta_{\text{DP}}$, it has the lowest accuracy at the same time, which means the $\Delta_{\text{DP}}$ is low due to the predictions are not good. We do not find obvious superiority for the fairness baselines compared with vanilla VAE when $\rho <0.2$. Overall, DBRF still works better w.r.t. effectiveness and robustness than other methods under different label bias settings in the multiple sensitive attributes case. 


\begin{wrapfigure}{R}{0.95\textwidth}
\centering
\includegraphics[width=0.95\textwidth]{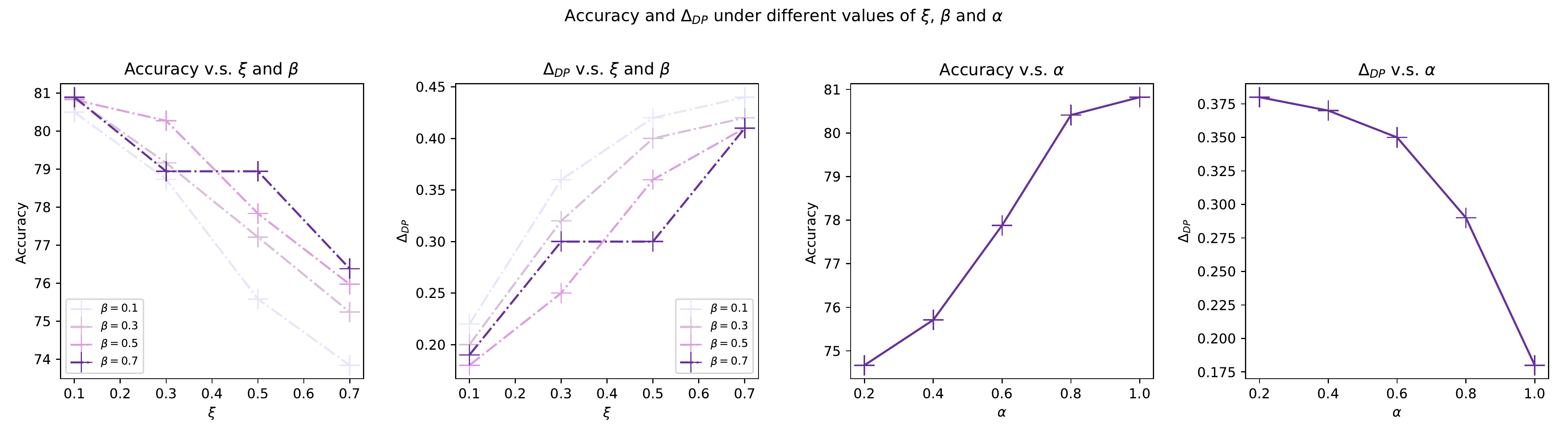}
\caption{Prediction performance and fairness measure on the Adult dataset with $\rho = 0.45$. }
\label{fig:hyper}
\end{wrapfigure}

\subsection{Hyperparameters}
We conduct experiments on different values of the hyperparameters $\lambda$, $\alpha$, $\beta$ and $\xi$. We first test different combinations of the two hyperparameters $\beta$ and $\xi$, which control the impact from $y$ on $b$ and $r_m$ respectively. We choose $\beta \in \{0.1,0.3,0.5,0.7\}$ and $\xi \in \{0.1,0.3,0.5,0.7\}$, while we fix $\alpha=1$ and $\lambda=0.1$. We visualize the relationship between the different combinations of $\beta$ and $\xi$ with accuracy and $\Delta_{\text{DP}}$ in \cref{fig:hyper}. We conduct the experiment with a large amount of label bias ($\rho=0.45$), so in such a scenario, if we increase the value of $\xi$ given the value of $\beta$ unchanged, this will guide the model to learn $r_m$ more from the biased label $y$. As a result, it causes the accuracy to drop and fairness violation to increase. On the contrary, if we increase the intensity of $\beta$ with a fixed value of $\xi$, the model will add more attention to correct $r_m$ from the unfair samples. Then, we fix $\beta = 0.5$, $\xi=0.1$ and $\lambda=0.1$ to test the impact of $\alpha$ on predictiveness. We plot different values of $\alpha$ from 0.2 to 1.0 against the accuracy and $\Delta_{\text{DP}}$ in \cref{fig:hyper}. The results show that as $\alpha$ increases, the predictiveness also increases with fairness violation decreasing. The trend is explained as when we increase the power of $\alpha$, $b$ will preserve more biased information, making $z$ fairer than the case when $\alpha$ is low. Moreover, this will impact the results from two aspects: the first is that the weight computed from $p(y\mid b)$ will be more confident; the second is that $r_m$ and $z$ will be more `clean'. Both aspects lead to the same outcome: improving performance and fairness. We also conduct experiments over different values for the hyperparameter $\lambda$ in the compression term. However, we do not observe any patterns that the change of $\lambda$ will affect the results (w.r.t. accuracy and fairness) too much. Overall, the results shown in \cref{fig:hyper} align with our expectation: $\alpha$ affects the predictiveness since it controls how much information $b$ contains from $a$, $\beta$ and $\xi$ control the influence from $b$ and $r_m$ respectively, and $\lambda$ does not influence the results too much since we have another KL term in \cref{eq:loss1_vae} to guarantee the disentanglement.

\subsection{Effectiveness of Objective Function}

\begin{wraptable}{r}{0.5\columnwidth}\vspace{-.45cm}
\caption{Ablation analysis.}
\centering
\resizebox{\linewidth}{!}{%
\begin{tabular}{c|c|c}
\toprule
& ACC (\%)   & $\Delta_{\text{DP}}$   \\ 
\midrule
$-\mathcal{L}_{\text{DBVAE}}+\mathcal{L}_{\text{UMI}}+ \mathcal{L}_{\text{p}}$      & 83.83$\pm$0.02 & 0.16$\pm$0.01      \\ 
\midrule
$-\mathcal{L}_{\text{DBVAE}}$ &  80.50$\pm$0.23 & 0.28$\pm$0.01  \\ 
\midrule
$-\mathcal{L}_{\text{DBVAE}}+\mathcal{L}_{\text{p}} $ & 79.89$\pm$0.09  & 0.05$\pm$0.01 \\ 
\midrule
$-\mathcal{L}_{\text{DBVAE}}+\mathcal{L}_{\text{UMI}}$ & 80.75$\pm$1.18    & 0.12$\pm$0.09 \\ 
\midrule
$-\mathcal{L}_{\text{DBVAE}}+\xi H(y\mid r_m)$      &79.22$\pm$1.48&    0.33$\pm$0.06  \\ 
\midrule
$-\mathcal{L}_{\text{DBVAE}}+\beta H(y\mid b)$      &  78.51$\pm$0.65     & 0.06$\pm$0.02     \\ 
\midrule
$-\mathcal{L}_{\text{DBVAE}}+\beta H(y\mid b)+\mathcal{L}_{\text{p}}$      &  79.21$\pm$0.16 & 0.06$\pm$0.01     \\ 
\bottomrule
\end{tabular}
}
\label{tab:ablation_analysis}
\vskip -0.0in
\end{wraptable}

\begin{figure}[t]
\centering
\includegraphics[width=\textwidth]{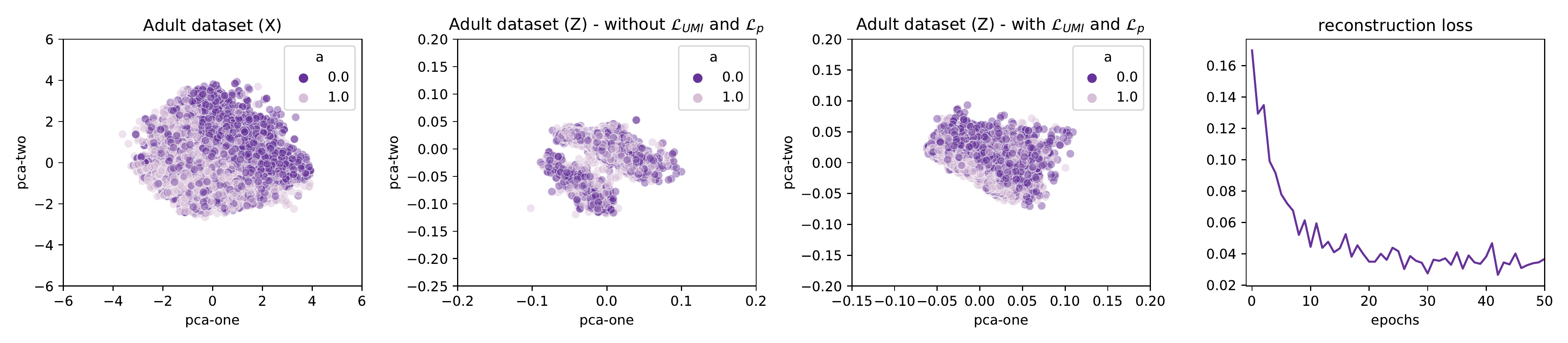}
\caption{Representation analysis. The first three plots depict the visualization of $x$, $z$ (without $\mathcal{L}_{\text{UMI}}$ and $\mathcal{L}_p$), $z$ (optimizing \cref{eq:objective_function} after applying kernel PCA. We use two colors to denote different demographic groups. The darker color denotes the protected group and the lighter color denotes the privileged group. The last plot is the convergence of the reconstruction loss.}
\label{fig:vis_representation}
\vskip -0.1in
\end{figure}

We use the Adult dataset to examine the effectiveness of the DBRF loss in \cref{eq:objective_function}. The results are shown in \cref{tab:ablation_analysis}. We compare the accuracy and fairness violation with different combinations of the components in \cref{eq:objective_function}. Based on the results, we can see that without the regularization terms $\mathcal{L}_{\text{UMI}}$ and  $\mathcal{L}_{\text{p}}$, we would have higher fairness violation, which aligns with our expectation. Then, if we remove the supervision loss $\mathcal{L}_{\text{p}}$ with only $\mathcal{L}_{\text{UMI}}$ left, we can achieve a lower fairness violation but with lower accuracy as well. Furthermore, the effects of term $H(y\mid r_m)$, $H(y\mid b)$ and the reconstruction of biased representations back to the sensitive attributes can be found in~\cref{fig:hyper}. We also conduct the analysis of the representations, which is shown in~\cref{fig:vis_representation},. We first plot the original $x$ after Kernel PCA w.r.t. two demographic groups. We can see the locations of data points in different demographic groups have a clear pattern, where the protected group locates in the upper right. Next, we visualize the representation $z$ using Kernel PCA w.r.t. the two demographic groups without regularization term $\mathcal{L}_{\text{UMI}}$ and supervision loss $\mathcal{L}_p$. From the results, we can see that without incorporating the label information, $z$ appears some randomness and still contains discriminative information. Then, we add the regularization term $\mathcal{L}_{\text{UMI}}$ and supervision loss $\mathcal{L}_p$ back, and the visualization result shows DBRF can successfully reduce the discrimination.

\section{Conclusions}
\label{conclusion}
In this work, we demonstrate that when the labels are unreliable, we can still learn fair representations and ideal labels by proposing DBRF, which utilizes the disentanglement VAE with the information bottleneck technique. We design DBRF based on two aspects: (1) disentangle the sensitive information from latent representations and (2) learn ideal labels by deriving the information bottleneck to discourage the ideal labels from being similar to observed biased ones for unfair samples. We derive a novel tractable objective function for optimizing DBRF. In experiments, we empirically demonstrate the superiority of DBRF to baseline models w.r.t. the effectiveness and robustness under different amounts of label bias. In addition, we also show that when the observed labels are unreliable, the learned fair representations are still discriminated against the particular demographic group since the prediction of downstream tasks is still measured with the unreliable labels. In DBRF, we rely on the prediction from $b$ to $y$ to measure how unfair the sample is. A possible research track in the future is to investigate more sophisticated measures instead of using the prediction confidence score.

\bibliographystyle{plain}
\bibliography{template}

\newpage
\appendix
\providecommand{\upGamma}{\Gamma}
\providecommand{\uppi}{\pi}
\section{Derivation of DBRF Objective Function}
\label{sec:app_mi_upper_bound}

By using the property of mutual information and applying chain rule, the $\mathcal{L}_{\text{IB}}$ term can be expanded as:
{\small
\begin{equation*}
\begin{aligned}
&\mathcal{L}_{\text{IB}} = \lambda I(x;z,b) + I(b;r_m) - \beta I(b;y) - \xi I(r_m;y)\\
&=\lambda (H(z,b) - H(z,b\mid x)) + I(b;r_m) - \beta I(b;y) - \xi I(r_m;y)\\
&=\lambda (H(z,b) + \int_{z,b,x}q(z,b,x)\log \frac{q(z,b,x)}{q(x)}dxdbdz)+ I(b;r_m) - \beta I(b;y) - \xi I(r_m;y)  \\
&=\lambda (H(z,b) + \int_{z,b,x}q(z,b,x)\log p(x\mid z,b) dxdbdz +\int_{z,b,x}q(z,b,x)\log \frac{q(z,b,x)}{p(z,b)q(x)}\frac{p(z,b)}{p(x\mid z,b)}dxdbdz)\\
&+ I(r_m;y) - I(r_m;y\mid b)+I(r_m;b\mid y) - \beta I(b;y) - \xi I(r_m;y) \\
&=\lambda (\mathbb{E}_{q(z,b\mid x)}\log p(x\mid z,b)+\text{KL}[q(z,b\mid x)||p(z,b)]-\int_{z,b}q(z,b\mid x)\log p(x\mid z,b)dbdz +H(z,b\mid x))\\
&+ I(r_m;y)+I(b;r_m \mid y)- \beta I(b;y) - \xi I(r_m;y)\\
&= \lambda (\mathbb{E}_{q(z,b\mid x)}\log p(x\mid z,b)+\text{KL}[q(z,b\mid x)||p(z,b)]-\int_{z,b}q(z,b\mid x)\log p(x\mid z,b)dbdz +H(z,b\mid x))\\
&+I(r_m;y)-H(y\mid b)+H(y\mid b,r_m) - \beta I(b;y) - \xi I(r_m;y)\\
&=\lambda (\mathbb{E}_{q(z,b\mid x)}\log p(x\mid z,b)+\text{KL}[q(z,b\mid x)||p(z,b)]-\int_{z,b}q(z,b\mid x)\log p(x\mid z,b)dbdz +H(z,b\mid x))\\
&+H(y) - H(y\mid r_m)-H(y\mid b)+H(y\mid b,r_m) - \beta (H(y) - H(y\mid b)) - \xi I(r_m;y)\\
&=\lambda \text{KL}[q(z,b\mid x)||p(z,b)]+(\xi-1) H(y\mid r_m)-(1-\beta) H(y\mid b)
  +H(y\mid b,r_m) + (1-\xi -\beta)H(y)\\
  &= \lambda \text{KL}[q(z,b\mid x)||p(z,b)]+\xi H(y\mid r_m) + \beta H(y\mid b)+(1-\xi -\beta)H(y) + \sum_{y}p(y)\log \frac{p(y\mid r_m)p(y\mid b)}{p(y\mid b,r_m)}\\
  &=\lambda \text{KL}[q(z,b\mid x)||p(z,b)]+\xi H(y\mid r_m) + \beta H(y\mid b) - (\xi+\beta)H(y)
\end{aligned}
\end{equation*}

Since $H(y)\ge 0$ and $\xi \in [0,1]$, $\beta \in [0,1]$, we rewrite $(\xi +\beta)H(y)$ as $C \ge 0$, and we define
\begin{equation*}
   \mathcal{L}_{\text{UMI}}:= \xi H(y\mid r_m) + \beta H(y\mid b) 
\end{equation*}}



\section{Implementation Details}
\label{sec:app_implementation}
We simply implement one layer encoder and decoder with the `ReLu' activation function. For the prediction classifier for $f(b)$ and $f(r_m)$, we implement a one hidden layer MLP with `ReLu' activation function and one dropout layer. For all the baseline models, we fix the encoder and decoder structure. For FFVAE, FVAE and VAE which requires downstream classifiers for the learned representations, we all use the same prediction classifier, which is the MLP with one hidden layer that we have used for $f(b)$ and $f(r_m)$.

For approximate $\text{KL}[q(z\mid x) || p(z)]$, we use the same method introduced in FactorVAE~\citep{pmlr-v80-kim18b}. We train a discriminator to distinguish the fake samples drawn from $p(z)p(b)=\mathcal{N}(0,I)\text{Uniform}(0,1)$ and `real' samples obtained from the encoder.Then, we can use 
\begin{equation*}
    \mathbb{E}_{q(z,b)}[\log d(u=1 \mid z,b)- 
    \log d(u=0\mid z,b)]
\end{equation*}
to approximate $\text{KL}[q(z\mid x) || p(z)]$, where $d$ is the discriminator, and we use $u=1$ to denote `real' samples and we use $u=0$ to denote `fake' samples. 

\section{Generation of Synthetic Data}
\label{sec:app_synthetic_data}

We generate two multivariate Gaussian distributions for each label class. For the positive class, we have $\mu = (2,2)$ and $\text{Cov}=[[5,1],[1,5]]$. For the negative class, we have $\mu = (-2,-2)$ and $\text{Cov}=[[10,1],[1,3]]$. 

Then we assign the sensitive attribute from a Bernoulli distribution where $p(a=1) = \frac{p(x'\mid y=1)}{p(x'\mid y=1)+p(x'\mid y=0)}$ and $x'$ is a transformed version of $x$ which can be computed by: $x' = \begin{bmatrix}
cos(\phi) & -sin(\phi)\\
sin(\phi)& cos(\phi)
\end{bmatrix}X$.

We totally generate 10800 synthetic samples with 2-dimensional non-sensitive attributes space and 1-d sensitive attribute space. We require the synthetic data to be as fair as possible. Therefore, the fairness violation on the synthetic data is very low, which is $\Delta_{\text{DP}} = 0.2$. 


\begin{wrapfigure}{R}{0.35\textwidth}
\vskip -0.2in
\centering
\includegraphics[width=0.35\textwidth]{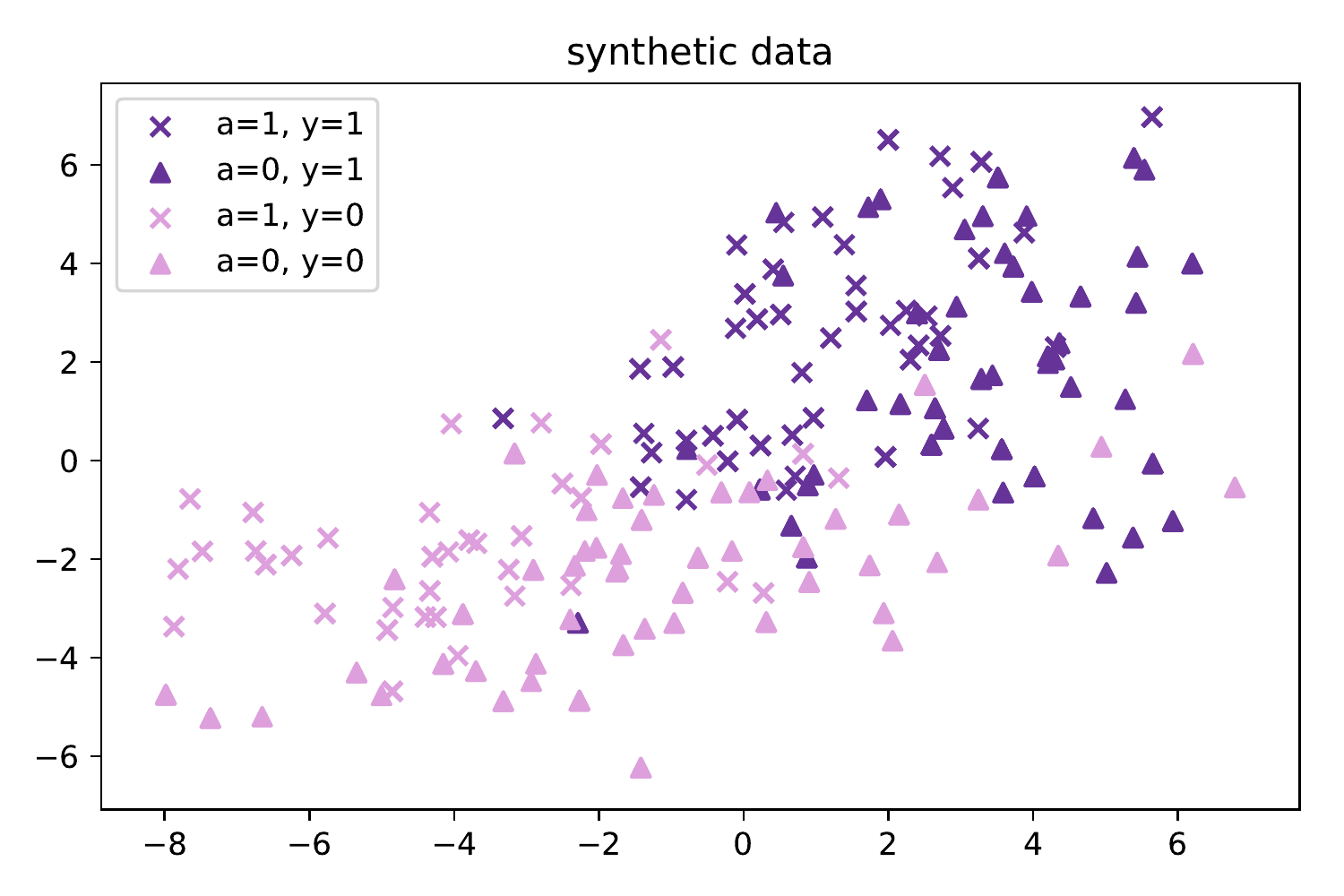}
\caption{Visualization for the synthetic data with 500 samples.}
\label{fig:syn} 
\vskip -0.1in
\end{wrapfigure}

\section{Datasets Description}
\label{sec:app_dataset}

\textbf{Synthetic Dataset: } We use the same synthetic data generation approach mentioned in fairness constraint method~\citep{2016_zafar}. We generate two multivariate Gaussian distributions for each label class. Then we randomly assign the sensitive attribute to each sample. For the synthetic data, we control the data to be fair by enforcing $\Delta_{\text{DP}}$ close to 0. For illustration, we only consider the binary sensitive attribute in synthetic data. 

\textbf{Adult Dataset\footnote{http://archive.ics.uci.edu/ml/datasets/Adult}:} The target value is whether an individual's annual income is over \$50k. The original feature dimension in this dataset is 13. After feature aggregation and encoding, the feature dimension is expanded to 35. The sensitive attributes are `Gender' and `Race'. In the binary sensitive attribute setting, we define `Gender' as our interested sensitive attribute and `Gender = Female' as the protected group. In the multi-attribute setting, we define `Gender' and `Race' as sensitive attributes and `Gender = Female' combined with `Race = Black' as the protected group. 

\textbf{Compas Dataset\footnote{ www.propublica.org/article/how-we-analyzed-the-compasrecidivism-algorithm}:} This data is from COMPAS, which is a tool used by judges, probation and parole officers to assess the risk of a criminal to re-offend. We focus on the prediction of `Risk of Recidivism' (Arrest). The Compas system is found to be biased in favor of white and female defendants over a two-year follow-up period. In the binary sensitive attribute setting, we define `Race' to be the target attribute and `Race = Black' as the protected group. In the multi-attributes setting, we define both `Race' and `Gender' as the sensitive attributes and `Race=Black' and `Gender=Male' as the protected group. After feature aggregation and encoding, the feature dimension is reduced to 11.

\section{Other Fairness Measure}
\label{sec:app_other_fair}
In this paper, we emphasize that our method does not need to specify the form of fairness notions in advance. We only report $\Delta_{\text{DP}}$ in \cref{sec:exp}, but our proposed method can be evaluated using other fairness notions. We also conduct experiments on difference of equal opportunity (DEO)~\citep{equal_opportunity}, which is defined as:
\begin{equation*}
    \text{DEO} = |\mathbb{E}(\hat{y}=1\mid y=1, a=1)-\mathbb{E}(\hat{y}=1\mid y=1,a=0)|.
\end{equation*}

Overall, the performance is similar as measured in $\Delta_{\text{DP}}$ while both DBRF* and DBRF+LR achieve lower fairness violations. For Compas Dataset, it is worth noting that, though FVAE has the lowest DEO, it has the lowest accuracy at the same time.

\begin{wrapfigure}{R}{0.85\textwidth}
\centering
\subfloat[Synthetic.\label{fig:syn_other}]{\includegraphics[width=0.3\textwidth]{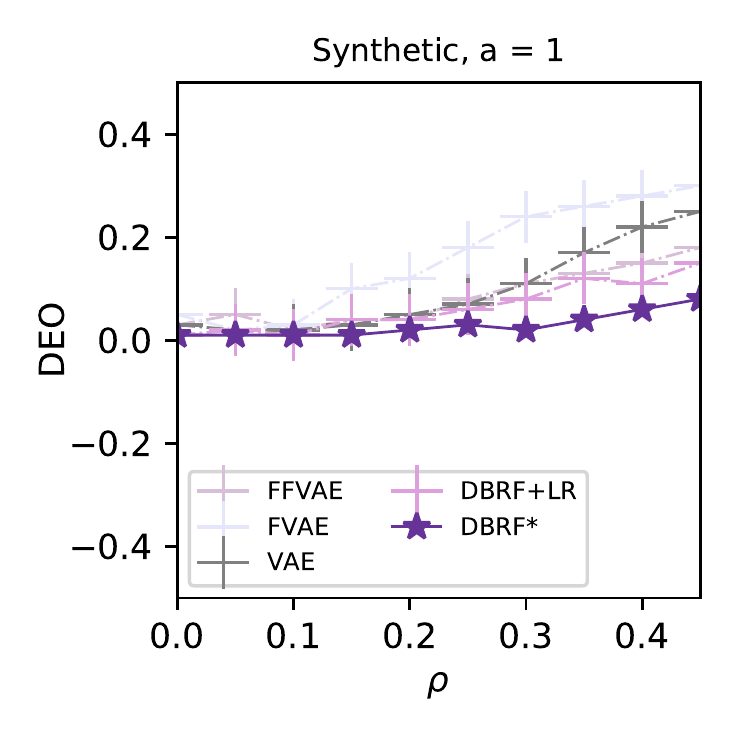}}
\subfloat[Adult.\label{fig:adult_other}] {\includegraphics[width=0.3\textwidth]{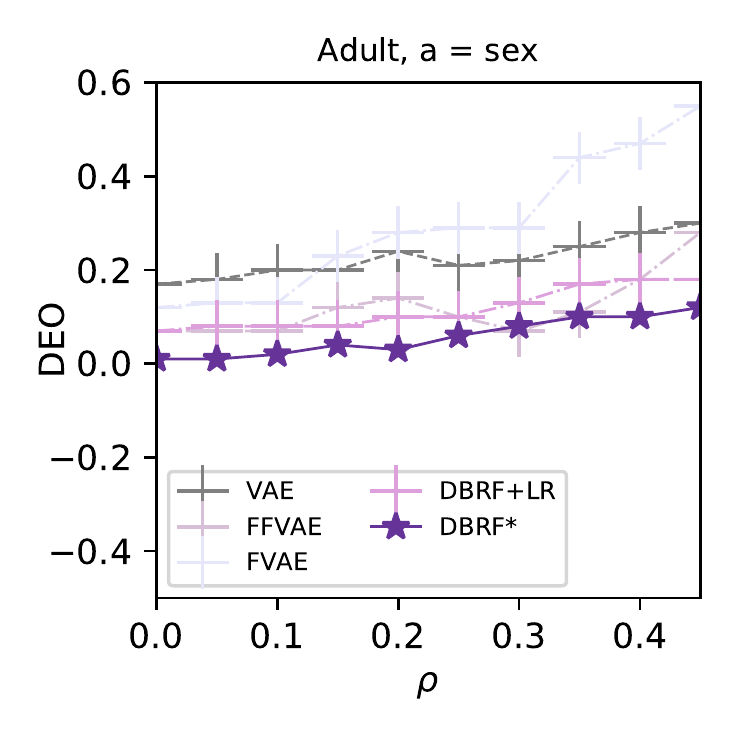}}
\subfloat[Compas.\label{fig:compas_other}]{\includegraphics[width=0.3\textwidth]{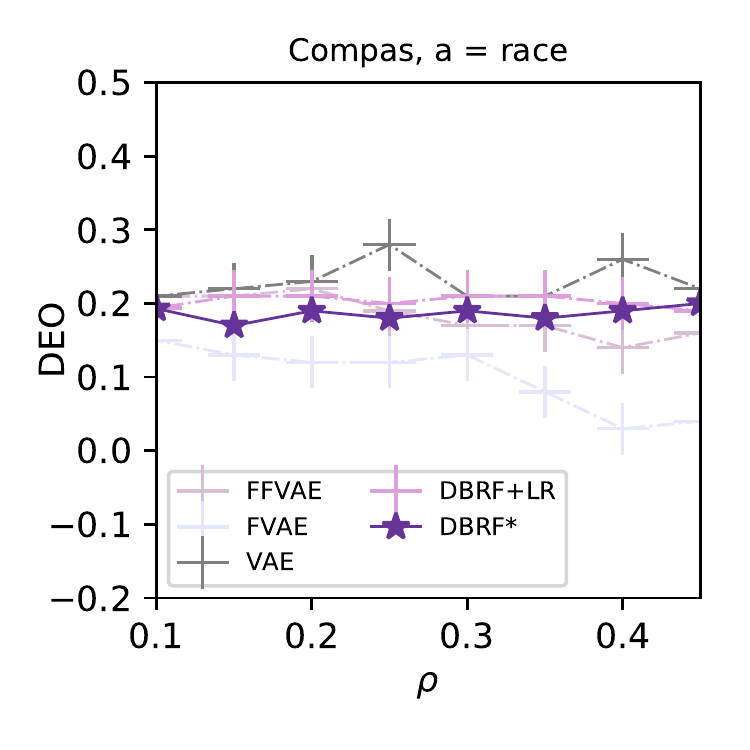}}
\caption{Accuracy and DEO under different label bias settings. We start from clean data and incrementally add the bias from 0.5 to 4.5. For DBRF, when $\rho<0.3$, we fix $\alpha=1$, $\lambda=0.1$, $\xi=0.1$, and $\beta=0.1$ for the reported results. When $\rho\geq0.35$, we increase $\beta$ to 0.5.}
\label{fig:deo_fairness}
\vskip -0.1in
\end{wrapfigure}

\end{document}